\documentclass[10pt,twocolumn,letterpaper]{article}

\usepackage[pagenumbers]{cvpr} 

\usepackage{amsmath}
\usepackage{algorithm}
\usepackage{algpseudocode}
\usepackage{xcolor}
\usepackage[normalem]{ulem}
\usepackage{xspace}
\usepackage{mathtools}
\usepackage{multirow}
\usepackage{subcaption}

\definecolor{cvprblue}{rgb}{0.21,0.49,0.74}
\usepackage[pagebackref,breaklinks,colorlinks,allcolors=cvprblue]{hyperref}

\def\paperID{4101} 
\def\confName{CVPR}
\def\confYear{2025}

\title{Learning from Streaming Video with Orthogonal Gradients}

\author{Tengda Han\textsuperscript{$\diamond$}, Dilara Gokay\textsuperscript{$\diamond$}, Joseph Heyward\textsuperscript{$\diamond$}, Chuhan Zhang\textsuperscript{$\diamond$} \\
Daniel Zoran\textsuperscript{$\diamond$}, Viorica P\u{a}tr\u{a}ucean\textsuperscript{$\diamond$}, Jo\~{a}o Carreira\textsuperscript{$\diamond$}, Dima Damen\textsuperscript{$\diamond\dagger$}, Andrew Zisserman\textsuperscript{$\diamond\ddag$}\\
{\small \textsuperscript{$\diamond$}Google DeepMind, \textsuperscript{$\dagger$}University of Bristol, \textsuperscript{$\ddag$}University of Oxford}
}

\begin{document}
\maketitle
\begin{abstract}
We address the challenge of representation learning from a continuous stream of video as input, in a self-supervised manner.
This differs from the standard approaches to video learning where videos are chopped and shuffled during training in order to create a non-redundant batch that satisfies the independently and identically distributed~(IID) sample assumption expected by conventional training paradigms.
When videos are only available as a continuous stream of input, the IID assumption is evidently broken, leading to poor performance.
We demonstrate the drop in performance when moving from \textbf{shuffled} to \textbf{sequential} learning on three tasks: the one-video representation learning method DoRA, standard VideoMAE on multi-video datasets, and the task of future video prediction.

To address this drop, we propose a geometric modification 
to standard optimizers, to decorrelate batches by utilising orthogonal gradients during training.
The proposed modification can be applied to any optimizer -- we demonstrate it with Stochastic Gradient Descent (SGD) and AdamW.
Our proposed orthogonal optimizer allows models trained from streaming videos to alleviate the drop in representation learning performance, as evaluated on downstream tasks.
On three scenarios (DoRA, VideoMAE, future prediction),
we show our orthogonal optimizer outperforms the strong AdamW in all three scenarios.
\end{abstract}    
\section{Introduction}
\label{sec:intro}

Trained on Internet-scale data at powerplant-scale energy costs, the way deep learning models are created today is drastically different from the way humans acquire their visual intelligence.
Humans perceive a single continuous visual input, starting from being infants in cribs.
This visual input is highly redundant and temporally correlated.
Such an input poses significant challenges for current deep learning paradigms.
These paradigms were primarily developed for learning from images, and make assumptions on the informativeness of every training batch as an independently and identically distributed (IID) sample from the data distribution.
These assumptions are immediately broken when learning from a continuous stream.

\begin{figure}[t]
    \centering
    \includegraphics[width=1\linewidth]{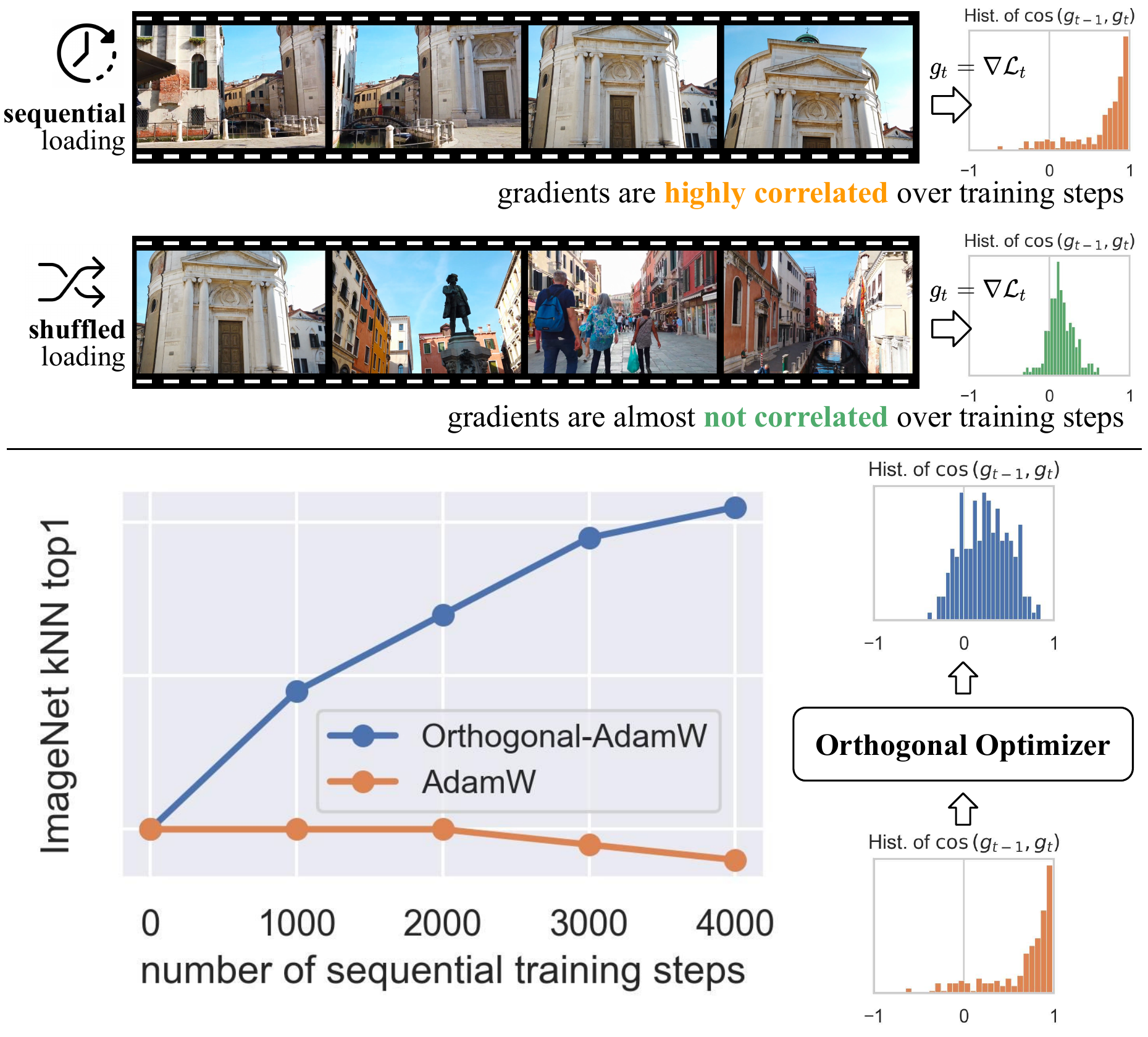}
    \caption{
    We address the task of learning from video by sequentially loading its clips in time (top).
    As neighbouring clips are very similar, consecutive gradients are highly correlated -- we show the histogram of cosine similarity of gradients between consecutive batches. This causes model collapse.
    In contrast, current methods shuffle the video to simulate an IID input (middle).
    Consecutive gradients are accordingly decorrelated -- cosine similarity is centred around 0.
    We propose to learn from the \textbf{orthogonal gradients} -- which allow standard optimizers to recover the drop in performance when training from a sequential video stream (bottom). 
    }
    \label{fig:teaser}
\end{figure}

To accommodate the current learning paradigms, video models have to date been restricted to learning from short clips, by dividing any long video streams into short segments and shuffling these to enable learning.
This gap between human learning and current video models is not only a computational burden from storing and accessing large videos, but is also potentially limiting the capabilities of models to achieve the human's ability to generalise.
Learning from a continuous stream is key to enabling intelligent agents that learn on-the-fly or adapt to new environments.
Additionally, models that learn from streaming videos can address privacy concerns as videos are not stored or shared.

Our paper is inspired by recent works that attempt to learn from a single video~\cite{venkataramanan2024dora} or from streams of videos~\cite{carreira2024bl}.
In~\cite{venkataramanan2024dora}, learning from a single long video is impressively shown to generalize but only when the video is stored in disk, such that random access is possible -- with batching, random sampling and shuffling.  
In contrast, \cite{carreira2024bl} concatenates videos to simulate a continuous stream that matches a day-long input and demonstrates the drop in performance when moving from shuffled to sequential learning on a number of self-supervised and supervised tasks.

In this work, we focus on the core obstacle to learning from streaming video: the redundancy of the data leading to highly correlated gradients. 
We make the following contributions:
\begin{itemize}
    \item We quantify the drop in performance when learning from sequential, rather than shuffled, data on three video learning methods: 
    DoRA, VideoMAE and future prediction.
    \item We propose to use a geometrically-principled optimizer, using the orthogonal gradient during learning.
    \item We augment commonly used optimizers -- Stochastic Gradient Descent (SGD) and AdamW with learning from orthogonal gradients. We refer to these augmented optimizers as orthogonal optimizers.
    \item We showcase clear improvements using our orthogonal optimizers when learning from sequential data on all three video learning methods.
\end{itemize}

\section{Related Work}
\label{sec:related_work}

\paragraph{Continual Learning.}
Existing continual learning literature focuses on the learning dynamics and the effects of introducing novel tasks over the training cycle, emphasizing knowledge accumulation as new tasks and data become available. The main objective of such works is quick adaptation while preserving performance on previously learned tasks (failure to do so is commonly termed as ``Catastrophic Forgetting")~\cite{kirkpatrick2017forget}.
In the context of continual learning, 
task changes over the training progress -- 
it could be a different objective function, 
or incremental annotated labels~\cite{buzzega2020darkexperiencegeneralcontinual, qu2024recentadvancescontinuallearning}. 
But often the data in these tasks consists of independent images ~\cite{lin2021clear,icarl}, which are much less correlated than consecutive video frames.

Various approaches have been proposed to tackle these problems: input replay buffers, which make the learning problem closer to the IID case by accumulating a dataset to sample from;  architectural adjustments ~\cite{Mai_2021_CVPR,ocl2019,onpro}, adapting the optimization algorithm~\cite{kirkpatrick2017forget}, or redesigning the training paradigm~\eg~adding pre-training with IID data~\cite{JMLR:v24:22-0496,ramasesh2022effect}.
Orthogonal gradients have been explored in the context of continual learning, where the model learns a number of distinct tasks iteratively~\cite{farajtabar2020ogd} -- in this case orthogonal gradients were used to avoid catastrophic forgetting of previously learned classes when learning continual image classification. The orthogonal computation is only computed after training each task. 

Different from prior work in continual learning, we address the problem of a \textit{single} task learnt from \textit{continuous} videos, where the learning process unfolds along the temporal dimension of visual sequences. 
This task is particularly challenging because in addition to the the problem of catastrophic forgetting due to the extensive temporal history, the continuity of the video frames introduces high correlations between consecutive learning steps which can be detrimental to the learning process.
We revisit~\cite{farajtabar2020ogd}, extending it to multiple optimizers and testing it for the first time on video tasks in general and streaming videos in particular.

\paragraph{Learning from Video Streams.}

The majority of video models trained today are trained from randomly sampled short clips sampled from large video datasets, creating roughly IID samples which make training with stochastic gradient descent effective. 

One exception is the work by Purushwalkam~\etal~\cite{purushwalkam2022continuous_ssl}, which explores learning a self-supervised model from continuous video streams. This work uses a `replay buffer' to store recent training samples in order to overcome the high temporal correlation of streamed videos.
Another work, closer to ours, is the `Baby Learning' framework~\cite{carreira2024bl}. In that work, a future prediction model is trained on streaming video and is evaluated on both in-stream and out-of-stream tasks -- trading off adaptation and generalization. This work includes experiments with a variety of common optimizers, but does not explicitly deal with the temporal correlation of the gradients.
Another line of work automatically filters training samples~\cite{aljundi2019gradient,evans2024curation} -- this can be applied to streamed video learning scenarios to handle the high temporal correlation of gradients. 
However, such methods effectively load more data than is actually used for training, and the result is quite similar to using a replay buffer, requiring extra compute and memory.

\paragraph{Video Representation Learning.}
Rapid progress has been made in visual representation learning from images and videos, especially in the family of self-supervised methods.
These can be grouped into three main core ideas -- contrastive learning (CPC~\cite{oord2018cpc}, MoCo~\cite{he2020moco}, SimCLR~\cite{chen2020simclr}, DPC~\cite{han2019dpc}), self-distillation (BYOL~\cite{grill2020byol}, DINO~\cite{caron2021dino}) and reconstruction based methods (MAE~\cite{he2022mae}, VideoMAE~\cite{tong2022videomae,feichtenhofer2022videomae}, SiameseMAE~\cite{gupta2023siamesemae}).
However, all these techniques rely on training with IID data randomly sampled from large, shuffled training datasets, which contrasts with the sequential nature in which humans, for example, perceive visual information. This paper explores some of these models when applied to the streaming video input scenario.

\paragraph{Test-time adaptation.}
When encountering distribution shifts at test time, models often fail to adapt or produce reasonable results given the new data. Test-time adaptation methods attempt to address this issue by either introducing training objectives which can be applied at test time~\cite{Liang2023ACS} - these usually would be self-supervised objectives~\cite{sun2019test,durasov20243} - or by adding regularization terms to an already trained model \cite{lin2023videotesttimeadaptationaction}. 
These ideas have been recently transposed to language models~\cite{sun2024learning} and connections to in-context learning and online reasoning are now actively being pursued~\cite{wang2023ttt, acurek2024TTT, durasov20243,openaio1, wang2024planningabilitiesopenaiso1}.
Here we demonstrate that test-time adaptation on video streams benefits from using orthogonal gradients. 

\section{Method}
\label{sec:method}

\paragraph{Problem Setup.}
We focus on the hard problem of sequential learning from a single continuous video.
Given a long video $\mathcal{V}$, 
our goal is to train a model $f_{\theta}$ on the video $\mathcal{V}$ \emph{sequentially} to minimize an objective function $\mathcal{L}$,
where $\theta$ represents the network parameters.
Since the entire video $\mathcal{V}$ is too long to feed into the model at once,
a practical approach is to cut the video into small chunks, $\{\mathcal{V}_1, \mathcal{V}_2, ..., \mathcal{V}_n\}$, where each $\mathcal{V}_i$ denotes $i$-th video clip with a short temporal window.
Different to the common practice of randomly sampling clips as mini batches to train the model,
we wish to learn from a video stream, so clips are fed in their \emph{sequential} order.

The greatest obstacles to learning from video sequentially is the 
high temporal correlation of gradients.
In most cases, the video changes slowly and the gradient of the current batch is almost identical to that of the previous batch.
The stochastic optimization methods widely used in deep learning, such as SGD, 
are all based on the assumption that the global gradient can be approximated by the gradient of mini-batches to some extent, which does not hold true when learning from sequential videos.
We focus on the task of learning from long videos in a \emph{self-supervised} manner where the learning signal purely comes from the pixels, and the temporal correlation of gradient is severe.
This is distinct from supervised learning where the supervisory signal might provide insights on where subtle changes or informative content is.
Our objective is a mechanism that can learn from these subtle changes; in effect, able to continually decorrelate the gradients and learn from the residual.

\paragraph{Learning from Orthogonal Gradients.}

\begin{figure}
    \centering
    \includegraphics[width=0.5\textwidth]{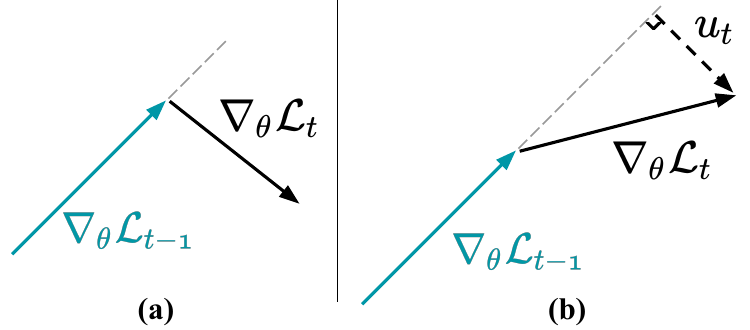}
    \caption{A simplified illustration of orthogonal gradients. 
(a) In common IID training, the gradient between consecutive steps are not very correlated due to the IID nature.
(b) Whereas if learning from sequential videos, the gradients between consecutive steps are highly correlated, which harms the optimization.
We propose to update the model parameters from the orthogonal components of the current gradient, denoted as $u_t$.
In practice, the gradients and the orthogonal operation are in a high dimensional space.
}
\label{fig:grad}
\end{figure}

Our method is straightforward: 
as the gradients are temporally correlated,
we propose to learn from the orthogonal components of the gradients.
In detail, 
the gradients of two consecutive update steps can be written as 
$g_{t-1} =  \nabla_{\theta}{\mathcal{L}_{t-1}}$ and 
$g_t = \nabla_{\theta}{\mathcal{L}_t}$, 
where $\theta$ denotes the model parameters and $\mathcal{L}$ is the loss function.
In an idealistic training scenario 
where the data samples in subsequent mini-batches follow the IID distribution, 
these two gradients typically have low similarity, 
which can be measured by a cosine distance $\cos{(g_{t-1}, g_{t})} \approx 0$.

When training sequentially, 
empirically we find the gradients between two consecutive update steps can be highly similar,~\ie~$\cos{(g_{t-1}, g_{t})} \rightarrow 1$, as shown in Figure~\ref{fig:teaser}.
To decorrelate these gradients, 
we propose to only update with the orthogonal component of the gradient $g_{t}$ w.r.t.\ the past gradient $g_{t-1}$ for the optimization step.
As illustrated in Figure~\ref{fig:grad},
the actual gradient used for the update is 
\begin{equation}
    u_t = g_{t} - \text{proj}_{g_{t-1}}{g_{t}}
\end{equation}
\noindent where $\text{proj}_{g_{t-1}}(g_{t})$ is the projection operation onto the direction $g_{t-1}$: 
\begin{equation}
\text{proj}_{g_{t-1}}(g_{t}) = \frac{g_{t}\cdot g_{t-1}}{g_{t-1}\cdot g_{t-1}} g_{t-1} = \frac{||g_{t}|| \cos{(g_{t}, g_{t-1})}}{||g_{t-1}||}  g_{t-1}
\label{eq:proj}
\end{equation}

This orthogonal gradient update has ideal behaviour for two scenarios at either end of the correlation spectrum:
(1) when the training data is close to an IID distribution,~\ie~$\cos{(g_{t-1}, g_{t})} \approx 0$,
the orthogonal gradient $u_t$ is close to the original gradient, since
$u_t = g_{t} - \text{proj}_{g_{t-1}}{g_{t}} \approx g_{t}$.
It means the orthogonal gradient based optimization rule is compatible with IID training scenario. 
In contrast, 
(2) when the consecutive data samples have high sequential similarity,~\ie~$\cos{(g_{t-1}, g_{t})} \approx 1$,
the orthogonal gradient has a small magnitude on a new direction 
$u_t = g_{t} - \text{proj}_{g_{t-1}}{g_{t}} \approx g_{t} - \frac{||g_{t}||}{||g_{t-1}||} g_{t-1} $. A small gradient results in minor changes to the model's parameters. 
This avoids the model to be excessively optimized along one gradient direction,
when there is insufficient new signal.

Practically, decorrelating the current gradient with the past \emph{single} step can be sensitive to noise.
Inspired by the common usage of `momentum' in standard optimizers~\cite{sutskever2013momentum},
we maintain an exponential moving average (EMA) of the 
original `clean' gradients, denoted by $c_{t}$, 
with an update rule 
\begin{equation}
    c_t \coloneqq \beta c_{t-1} + (1-\beta) g_t
\end{equation}
\noindent where $\beta$ is the momentum factor, by default we use $\beta=0.9$.
The orthogonal gradient is then computed by 
\begin{equation}
u_t = g_{t} - \text{proj}_{{c_{t-1}}}{g_{t}}
\end{equation}
Notice that the EMA is computed on the original gradients, rather than the orthogonal component $u_t$, whereas $u_t$ can be further fed into first/second order moment subject to the choices of optimizers~(\eg~second-order optimizer AdamW in Algorithm~\ref{algo:adamw}). 

Importantly, the aforementioned geometric modification is applicable to many optimizers.
Here we show two commonly used optimizer algorithms modified by orthogonal gradients: SGD optimizer as an illustration (Algorithm~\ref{algo:sgd}),
and the AdamW optimizer~\cite{loshchilov2017adamw} (Algorithm~\ref{algo:adamw}).
The\colorbox{green!20}{text in green}indicates the addition to the original algorithms.
We mostly experiment with Orthogonal-AdamW due to its faster convergence speed.

\begin{algorithm}[t]
\caption{\colorbox{green!20}{Orthogonal} SGD}

\begin{algorithmic}[1]
\Require Learning rate $\eta > 0$, momentum parameter $\beta \in [0, 1)$, initial parameters $\theta_0$, number of iterations $T$
\State \colorbox{green!20}{Initialize velocity $c_0 = 0$}
\For{$t = 1$ to $T$}
    \State Sample a mini-batch of data $\mathcal{B}_t$ from the training set
    \State Compute the gradient: $g_t = \nabla_{\theta} \mathcal{L}(\theta_{t-1}; \mathcal{B}_t)$
   \State \colorbox{green!20}{\resizebox{0.9\linewidth}{!}{Compute the orthogonal gradient: $u_t = g_t - \text{proj}_{c_{t-1}}{g_t}$}}
    \State \colorbox{green!20}{\resizebox{0.85\linewidth}{!}{Update the raw momentum: $c_t = \beta c_{t-1} + (1 - \beta) g_t$}}
    \State \colorbox{green!20}{{Overwrite the gradient: $g_t := u_t$}}
    \State Update the parameters: $\theta_t = \theta_{t-1} - \eta g_t$
\EndFor
\end{algorithmic}
\label{algo:sgd}
\end{algorithm}

\begin{algorithm}[t]
\caption{\colorbox{green!20}{Orthogonal} AdamW}
\begin{algorithmic}[1]
\Require Learning rate $\eta > 0$, weight decay coefficient $\lambda > 0$, 
         decay rates \colorbox{green!20}{$\beta,$}$\beta_1, \beta_2 \in [0, 1)$, small constant $\epsilon > 0$,
         initial parameters $\theta_0$, number of iterations $T$
\State Initialize first moment vector $m_0 = 0$, \colorbox{green!20}{$c_0 = 0$}, and second moment vector $v_0 = 0$
\For{$t = 1$ to $T$}
    \State Sample a mini-batch of data $\mathcal{B}_t$ from the training set
    \State Compute the gradient: $g_t = \nabla_{\theta} \mathcal{L}(\theta_{t-1}; \mathcal{B}_t)$
    \State \colorbox{green!20}{\resizebox{0.9\linewidth}{!}{Compute the orthogonal gradient: $u_t = g_t - \text{proj}_{c_{t-1}}{g_t}$}}
    \State \colorbox{green!20}{\resizebox{0.85\linewidth}{!}{Update the raw momentum: $c_t = \beta c_{t-1} + (1 - \beta) g_t$}}
    \State \colorbox{green!20}{{Overwrite the gradient $g_t \coloneqq u_t$}}
    \State Update biased first moment estimate: $m_t = \beta_1 m_{t-1} + (1 - \beta_1) g_t$
    \State Update biased second moment estimate: $v_t = \beta_2 v_{t-1} + (1 - \beta_2) g_t^2$
    \State Compute bias-corrected first moment: $\hat{m}_t = \frac{m_t}{1 - \beta_1^t}$
    \State Compute bias-corrected second moment: $\hat{v}_t = \frac{v_t}{1 - \beta_2^t}$
    \State Apply weight decay: $\theta_{t-1} = \theta_{t-1} - \eta \lambda \theta_{t-1}$
    \State Update parameters: $\theta_t = \theta_{t-1} - \eta \frac{\hat{m}_t}{\sqrt{\hat{v}_t} + \epsilon}$
\EndFor
\end{algorithmic}
\label{algo:adamw}
\end{algorithm}

\paragraph{Trade-off between algorithm and speed.}
In the convex optimization literature, 
there are relevant methods that might be more favourable than orthogonal operation, 
such as conjugate gradient method~\cite{shewchuk1994conjugate,hestenes1952conjugate}.
However, orthogonal gradient is computationally cheaper than conjugation, since the orthogonal projection can be implemented as cosine distance and vector norms (Equation~\ref{eq:proj}),
which could take advantages from well-optimized pre-compiled kernels in deep learning toolboxes.
We do not delve into this direction in this paper, but it could be an interesting future work.

\section{Experiments}
\label{sec:exp}

In this section, 
we focus on providing empirical evidence using real video datasets to demonstrate the effectiveness of the orthogonal optimizer. 
We particularly experiment with Orthogonal-AdamW, on three scenarios: 
representation learning on a {\em single} long video,
representation learning on video datasets, 
and future prediction tasks as same as~\cite{carreira2024bl}.

\begin{table*}[t]
\setlength{\tabcolsep}{3pt}
\small
\centering
\begin{tabular}{lll|ll}
\toprule
\multirow{2}{*}{initialization} & \multicolumn{2}{c|}{pretraining dataset: $\text{WT}_{\text{venice}}$}           & \multicolumn{2}{c}{downstream ImageNet} \\\cline{2-5}
& pretraining method              & optimizer        & linear probe top1 & kNN top1              \\ \hline
\color{gray}$\text{DINO}_\text{ImageNet}$ & \color{gray}- & \color{gray}- & \color{gray}-  & \color{gray}74.4 \\
$\text{DINO}_\text{ImageNet}$ & DoRA sequential (batch-along-time) & AdamW              & 6.1            & 1.8             \\
$\text{DINO}_\text{ImageNet}$ & DoRA sequential (batch-along-time) & Orthogonal-AdamW   & \textbf{64.5}           & \textbf{51.8}             \\ \hline
\color{gray}$\text{VideoMAE}_\text{SSV2}$ & \color{gray}- & \color{gray}- & \color{gray}- & \color{gray}3.7               \\
$\text{VideoMAE}_\text{SSV2}$ & DoRA sequential (batch-along-time) & AdamW              & 7.9  & 3.0  \\
$\text{VideoMAE}_\text{SSV2}$ & DoRA sequential (batch-along-time) & Orthogonal-AdamW   & \textbf{11.2}  & \textbf{5.7}   \\ 
\hline
random & DoRA sequential (batch-along-time) & AdamW              & 3.5  & {0.8}  \\
random & DoRA sequential (batch-along-time) & Orthogonal-AdamW   & \textbf{8.2}  & {\textbf{3.1}}   \\ 
\bottomrule
\end{tabular}\
\vspace{-2mm}
\caption{Experiments on DoRA~\cite{venkataramanan2024dora} pretraining on $\text{WT}_{\text{venice}}$}
\label{table:dora-venice}
\vspace{-2mm}
\end{table*}

\begin{figure*}[h!]
    \centering
    \small
    \includegraphics[width=0.95\textwidth]{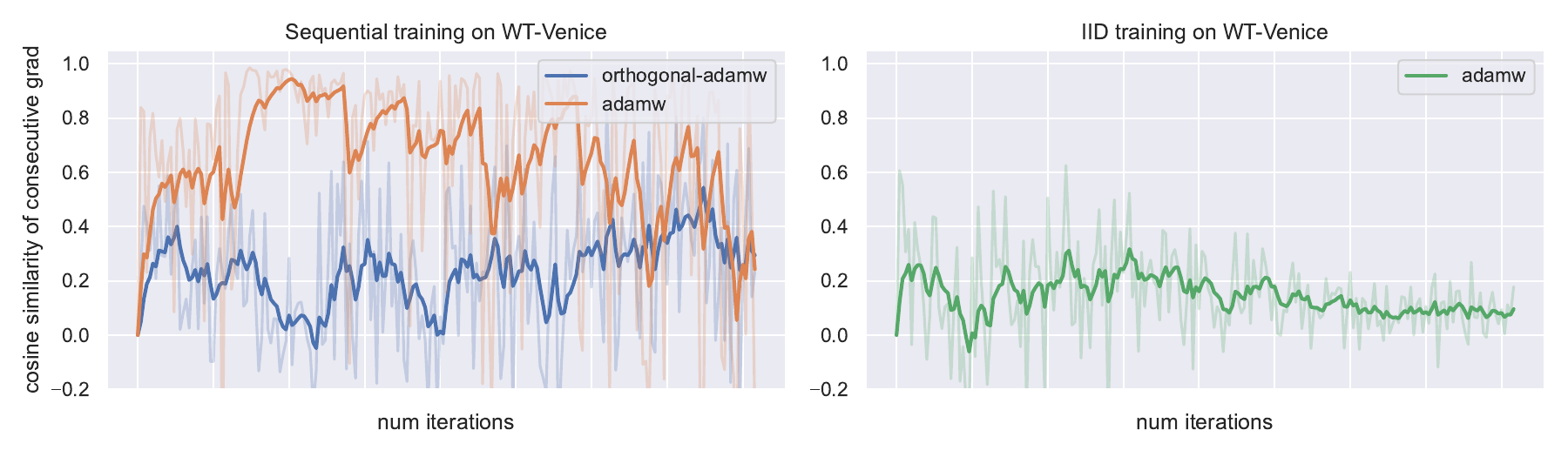} 
    \vspace{-3mm}
    \caption{Effect of orthogonal optimizer on sequential training of DoRA on the $\text{WT}_{\text{Venice}}$ video.
    On IID training, the consecutive gradient has low cosine similarity (right). 
    Sequential training (left) naturally brings a high similarity of consecutive gradient, 
    but the orthogonal optimizer decorrelate the gradients over time.
    Notice that we plot $\cos{(g_{t-1}, g_t)}$ in this figure.
    }
    \vspace{-2mm}
    \label{fig:cos_progress}
\end{figure*}

\subsection{DoRA on a Single Video}
The DoRA paper~\cite{venkataramanan2024dora} trains a vision transformer~\cite{dosovitskiy2020vit} image backbone on a single long video and achieves competitive performance. They apply aggressive frame augmentations and randomly sample short video clips, similar to other self-supervised works~\cite{caron2021dino,he2020moco}, to obtain diverse training samples.
Differently, we focus on learning video representation from a single video in a \emph{sequential} manner, 
which poses a great challenge to these prior methods because of the high temporal correlation between batches.

\paragraph{Datasets.}
Following DoRA~\cite{venkataramanan2024dora},
we use the WalkingTour video at Venice (denoted as $\text{WT}_{\text{venice}}$) from the \textbf{WalkingTour} dataset proposed by the same work.
This video is extensively used by DoRA and enables us to conduct a through analysis.
The $\text{WT}_{\text{venice}}$ video has a duration of 1 hour 50 minutes at 60fps, containing a continuous urban view around Venice city center filmed from a hand-held camera.
As in~\cite{venkataramanan2024dora}, we train on this video, albeit in a sequential manner, and then evaluate the learnt representation on the downstream task of object recognition on ImageNet~\cite{deng2009imagenet}.

\paragraph{Task Setting.}
We train the DoRA method using their official codebase.
Instead of randomly sampling short video clips,
we sample clips sequentially from the beginning of the video.
~\ie~given a batch size $N$, our first batch $\mathcal{B}_1 = \{ {C_1}, \dots, {C_N} \}$ 
and the second batch continues from 
${C_{N+1}}$ to ${C_{2N}}$, 
where $C_i$ denotes the $i$-th short clip from the source video.
Practically, each video clip contains 8 consecutive video frames sampled at 1 fps,
and every two consecutive clips are shifted by 1 frame, or $1/60$ second.
For evaluation, we monitor the performance of ImageNet linear probe 
and k-nearest-neighbour classification performance, same as DoRA.
We monitor the performance drop due to sequential video training, 
and observe how much gain the orthogonal optimizer can reclaim.

\paragraph{Architecture.}
We use the same architecture as DoRA, 
which consists of two ViT-S image backbones, 
forming a teacher-student structure. 
Each ViT-S backbone contains 12 transformer blocks with 384 embedding dimension.
Their training scheme is inspired by DINO~\cite{caron2021dino} -- 
the `teacher' module is updated from an exponential moving average~(EMA) of the student's parameters.
The DoRA architecture also contains a multi-object tracking module, which masks out objects based on the attention scores among the image patches produced by the teacher module. 
The teacher module has the privilege to observe the full video frames as input,
whereas the student module takes as input either heavily cropped video frames, or partially masked frames, and is trained to produce a representation that is close to the teacher's output;
the student module is trained with gradient back propagation.
For evaluation, we take the backbone of the teacher module and perform downstream tasks, 
same as DoRA.

\paragraph{Implementation Details.}
The original DoRA is trained from scratch for a long time (10+ days on 16 GPUs).
We train DoRA with different initialization methods including DINO weights pretrained on ImageNet,
and VideoMAE weights pretrained on Something-Something-V2 (SSV2)~\cite{goyal2017something}, and random initialization.
By default,
we train DoRA for 1 epoch on the $\text{WT}_{\text{venice}}$ 
with a batch size of 32 video clips distributed on 4 Nvidia A100 GPUs.
For ImageNet classification and kNN evaluation, we use the same setting as DoRA's codebase.
The full implementation detail can be found in the appendix.

\paragraph{DoRA Discussion.}
Figure~\ref{fig:cos_progress} illustrates the cosine similarity between consecutive gradients (\ie~$\cos{(g_{t-1}, g_{t})}$)
when training DoRA on $\text{WT}_{\text{venice}}$. 
It is clear that in sequential training scenario, 
the proposed Orthogonal-AdamW is able to reduce the gradient correlation over time (orange vs. blue), getting closer to the low correlation in IID sampling scenario (green).

The experimental results of training DoRA sequentially 
are shown in Table~\ref{table:dora-venice}.
When initializing with a strong $\text{DINO}_\text{ImageNet}$ checkpoint,
the Orthogonal-AdamW optimizer is able to prevent the training failure;
whereas with the baseline AdamW optimizer, 
the model parameters are damaged by the sequential training and cannot be trained further.
With $\text{VideoMAE}_\text{SSV2}$ initialization,
in a short training schedule the Orthogonal-AdamW optimizer surpasses AdamW on the same setting (3.0 to 5.7 on kNN accuracy).
We note that $\text{VideoMAE}_\text{SSV2}$ initialization gives much worse results on downstream ImageNet classification performance. This is possibly because the SSV2 dataset does not have enough diversity for general objects.
We also experimented training DoRA from scratch, although the sequential training of DoRA is inefficient, the Orthogonal-AdamW outperforms AdamW by a clear margin, and with AdamW the model does not train.

\begin{figure}[t!]
    \centering
    \includegraphics[width=0.45\textwidth]{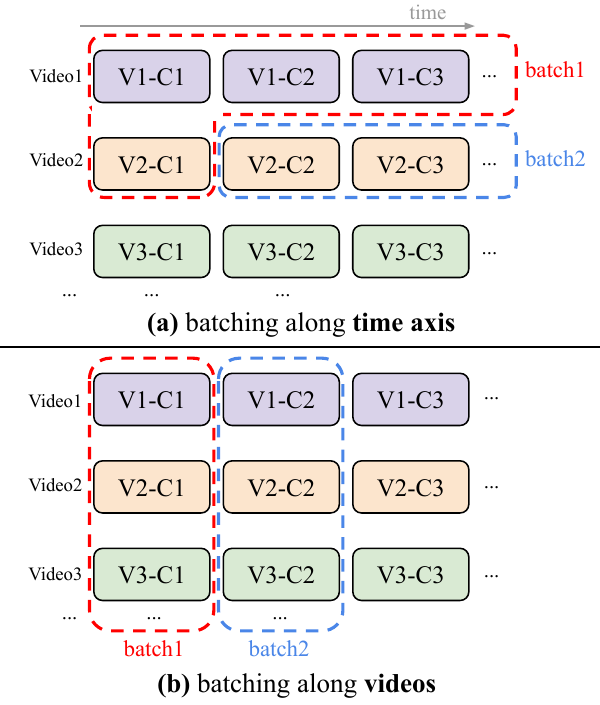}
    \vspace{-2mm}
    \caption{Two batch strategies for sequential video datasets, for videos $\text{V}_i$ divided into clips $\{\text{C}_1, ... \text{C}_N\}$. 
    \textbf{(a) batch along the time axis}: a more practical way of batching long video streams, where the samples within a batch have high correlation. But when the batch size is large, the temporal correlation between consecutive batches might be low.
    \textbf{(b) batch along videos}: samples within a batch are diverse but the temporal correlation between consecutive batches is high.
    Notice that in practice adjacent clips may have temporal overlaps,
    for clarity we do not show any overlaps in the figure. 
    }
    \vspace{-3mm}
    \label{fig:batch}
\end{figure}

\begin{table*}[h]
\setlength{\tabcolsep}{3pt}
\centering
\small

\parbox{\linewidth}{\centering
\begin{tabular}{ll|ll}
\toprule
\multicolumn{2}{c|}{pretraining dataset: SSV2}           & \multicolumn{2}{c}{downstream SSV2} \\ \hline
video processing       & optimizer         & linear-probe top1$\uparrow$ & attn-probe top1$\uparrow$ \\ \hline
\color{gray}$\text{VideoMAE}_\text{SSV2}$~\cite{tong2022videomae}   & \color{gray}AdamW  & \color{gray}23.2 &  \color{gray}55.7     \\ \hline
shuffled video clips   & AdamW                    & 19.0           &  \textbf{54.9}     \\
shuffled video clips   & Orthogonal-AdamW         & \textbf{21.0}  & {54.7}           \\ \hline
sequential (batch-along-time) & AdamW             & 16.4    &  46.1         \\
sequential (batch-along-time) & Orthogonal-AdamW  & \textbf{18.4} & \textbf{48.0}     \\ \hline
sequential (batch-along-video) & AdamW            & {9.5}  &  30.3           \\
sequential (batch-along-video) & Orthogonal-AdamW & {\textbf{10.4}} & \textbf{32.6}   \\ \bottomrule
\end{tabular}
}

\vspace{1em}

\parbox{\linewidth}{\centering
\begin{tabular}{ll|ll}
\toprule
\multicolumn{2}{c|}{pretraining dataset: K400}           & \multicolumn{2}{c}{downstream SSV2} \\ \hline
video processing       & optimizer       & linear-probe top1$\uparrow$ & attn-probe top1$\uparrow$   \\ \hline
\color{gray}$\text{VideoMAE}_\text{K400}$~\cite{tong2022videomae}   & \color{gray}AdamW   & \color{gray}19.2  & \color{gray}52.1 \\ \hline
shuffled video clips   & AdamW                      & 20.3  & 46.3              \\ 
shuffled video clips   & Orthogonal-AdamW  & \textbf{21.4}  & \textbf{48.4}   \\ \hline
sequential (batch-along-time) & AdamW           & 19.3           & 44.7             \\
sequential (batch-along-time) & Orthogonal-AdamW & \textbf{20.5}  & \textbf{46.5}    \\ \hline
sequential (batch-along-video) & AdamW           & \textbf{18.7}  & 43.5              \\
sequential (batch-along-video) & Orthogonal-AdamW & 18.2  & \textbf{43.6}             \\
\bottomrule
\end{tabular}
}
\vspace{-1mm}
\caption{Experiments on VideoMAE pretraining on SSV2 and K400.
The experiment in {\color{gray}gray} is our downstream evaluation results with the official checkpoint obtained from~\cite{tong2022videomae}.
}
\label{table:videomae_ssv2}
\end{table*}

\subsection{VideoMAE on Video Datasets}
For general self-supervised video representation learning, VideoMAE~\cite{tong2022videomae,feichtenhofer2022videomae} remains a competitive method which learns from reconstructing video patches, but it was mostly applied on large scale video datasets with large diversity. 
In this section, we generalize the proposed orthogonal optimizer to VideoMAE training on common video datasets rather than a single video, but in a sequential manner.

\paragraph{Batching strategy for sequential videos.}

From one video, loading clips in a sequential manner is straightforward.
But when the dataset has multiple videos, 
or there are multiple video streams available simultaneously,
two different ways of forming mini-batches emerge.
As shown in Figure~\ref{fig:batch} (a),
one can batch video clips over the time axis, and go through videos one by one in the dataset, such as $\mathcal{B}_1 = \{{V_1 C_1}, {V_1 C_2}, \dots\}$. But if the video in the dataset is not long enough w.r.t.\ the batch size,
the next batch might sample clips from a different video source (not from ${V_1}$).
As shown in Figure~\ref{fig:batch} (b),
one can also batch video clips over different videos,~\eg~$\mathcal{B}_1 = \{{V_1 C_1}, {V_2 C_1}, \dots\}$, and the next batch will sample videos from the next timestamp. 
In this section, we experiment with both batching methods, named as `batch-along-time' and `batch-along-video'.

\paragraph{Datasets.}
We use Something-Something-V2 and Kinetics\-{}-400 as pretraining datasets, to be compariable with VideoMAE~\cite{tong2022videomae}.
\textbf{Something-Something-V2~(SSV2)}~\cite{goyal2017something} is a fine-grained action classification dataset focusing on object manipulation. It consists of 220k short videos with duration between 2 to 6 seconds, which are labelled into 174 classes emphasising the action, such as `moving something from left to right'.
\textbf{Kinetics-400 (K400)}~\cite{kay2017kinetics} is a large scale action classification dataset sourced from internet videos. It contains 230k videos with duration of 10 seconds, spanning over 400 general human action classes.
For downstream evaluation, we report action classification results on SSV2 dataset.

\paragraph{Task Setting.}
We pretrain a VideoMAE model \emph{from scratch}, on both SSV2 and K400 datasets.
Differently to common practice that ramdomly samples short video clips from each video in the dataset and applies shuffling, we load videos in a sequential manner with both batching strategies illustrated in Figure~\ref{fig:batch}.
To evaluate the quality of learned representation, 
we apply two methods: linear-probe and attn-probe.
\textbf{Linear-probe} means a single linear layer on top of the frozen pre-trained visual encoder is trained for the action classification task; 
\textbf{Attn-probe} means attentive probing used in~\cite{bardes2024vjepa}: a single transformer block including attention operation and MLP layers is trained on top of the frozen pre-trained visual encoder for action classification task.

\paragraph{Architecture.}
We use a Vision Transformer~\cite{dosovitskiy2020vit} ViT-B as the visual encoder,
which consists of 12 transformer blocks with an embedding dimension of 768.
As part of the MAE training, we use a visual decoder which consists of 4 transformer blocks, 
which is trained to reconstruct visual patches from the encoder outputs, 
and will be discarded when evaluating for downstream tasks.
We train the VideoMAE with the default 0.9 drop ratio, which means 90\% of the visual patches will be discarded for the visual encoder and will be reconstructed by the visual decoder.
The entire network is trained from scratch.

\paragraph{Implementation Details.}
The model takes 16 frames at $224\times224$ resolution as input.
For sequential loading, 
we first take all the frames from each video, then sample 16-frame clips from that.
In order to have a clear comparison for both batching methods,
we sample the same number of clips from each video. 
For example, on SSV2 we first take 64 uniformly-sampled frames from each video, 
then take 4 clips without overlap, each clip containing 16 frames;
Similarly, on K400 we first take 112 uniformly-sampled frames from each video, 
then take 7 clips without overlap, each clip containing 16 frames.
With this, the models trained with both strategies observe exact same video clips, but only different in the batch arrangement.
For IID sampling and downstream tasks,
we use the default strategy as in~\cite{tong2022videomae}, 
where a 16-frame clip is randomly sampled from each video.
For our pretraining experiments,
the model is trained with a batch size of 512 clips for the same number of iterations (260k steps),
for a fair comparison.
Other implementation details are in the appendix.

\begin{table*}[t!]
\setlength{\tabcolsep}{3pt}
\centering
\small
\resizebox{\linewidth}{!}{
\begin{tabular}{lll|ll|ll}
\toprule
displacement \textbf{+0.64s}                   & & & \multicolumn{2}{|l}{Ego4D: Pixel MSE$\downarrow$ / PSNR$\uparrow$} & \multicolumn{2}{|l}{ScanNet: Pixel MSE$\downarrow$ / PSNR$\uparrow$}  \\ \hline
method (batch-along-time)          & pretraining              & optimizer & in-s.   & out-of-s.      & in-s.  & out-of-s.       \\ \hline
BabyLearning~\cite{carreira2024bl} & Guided Future Prediction & RMSProp   & 0.055 / -   & 0.066 / -          & 0.055 / -  & 0.061 / -  \\ 
BabyLearning~\cite{carreira2024bl}$\dagger$ & ViT-L-I21K-CLS           & RMSProp   & 0.059 / -   & 0.073 / -          & 0.061 / -  & 0.066 / -  \\ \hline
BabyLearning (repro)               & ViT-L-I21K-CLS           & RMSProp   & 0.032 / 15.9  & 0.026 / 16.9          & 0.033 / 15.07    & 0.041 / 14.28      \\
BabyLearning                       & ViT-L-I21K-CLS           & AdamW     & 0.034 / 15.9  & 0.026 / 16.8          & 0.033 / 15.72    & 0.033 / 15.27        \\
BabyLearning                       & ViT-L-I21K-CLS   & Orthogonal-AdamW  & \textbf{0.031} / \textbf{16.4}  & \textbf{0.023} / \textbf{17.6} & \textbf{0.032} / \textbf{15.77} & \textbf{0.033} / \textbf{15.28}  \\
\bottomrule
\end{tabular}}
\vspace{-2mm}
\caption{Performance on future frame prediction task on Ego4D-Stream and ScanNet-Stream datasets, compared with~\cite{carreira2024bl}.
$\dagger$ this result are obtained by contacting the authors.
The `(repro)' denotes our reproduction of the experiment from~\cite{carreira2024bl} with a same setting.
}\label{table:blv1}
\end{table*}

\paragraph{Discussion.}
The results are shown in Table~\ref{table:videomae_ssv2}.
First, notice that there is no big drop when switching from IID sampling to the `batch-along-time' sequential sampling,~\eg~linear probe $19.0\rightarrow16.4$ for SSV2, $20.3\rightarrow19.3$ for K400.
The reason is the videos in SSV2 and K400 are relatively short compared with our batch size (512 clips), the consecutive batches actually contain clips from different video sources.
Second, it is expected that `batch-along-video' gives worse results than `batch-along-time' due to larger temporal correlation between batches.
Third, proposed Orthogonal-AdamW optimizer works better than the baseline AdamW on both sequential cases,~\eg~attn probe top1 +2\% when pretrained on SSV2, and +1\% when pretrained on K400.
Additionally, it is interesting that the Orthogonal-AdamW also works slightly better on shuffled clips,~\eg~linear probe top1 +2\% when pretrained on SSV2 and +1\% when pretrained on K400. 
Probably it is because the inter-batch correlations from shuffled clips on SSV2 and K400 datasets are significant enough, that decorrelating the gradients bring some small gains.

\subsection{Future Prediction on Video Streams}
In this section, we reproduce the experiments of learning from video streams from Carreira~\etal~\cite{carreira2024bl},
and experiment with our orthogonal optimizer on this sequential training scenario.

\paragraph{Datasets.}
Following~\cite{carreira2024bl}, we use ScanNet-Stream and Ego4D-Stream datasets.
\textbf{ScanNet-Stream} is a continuous version of ScanNet-V2 proposed in~\cite{carreira2024bl}, which simply stitches all the videos together to mimic a long video and to experiment with sequential loading. ScanNet-V2~\cite{dai2017scannet} contains videos of in-door room scanning scenario, with an average duration of 1 minute, together with synchronized depth masks, semantic segmentation masks, and camera poses. We use the same train-val split as~\cite{carreira2024bl} -- 1.2k original ScanNet videos for training and 312 for validation. 
Similarly \textbf{Ego4D-Stream} is a stitched version of Ego4D~\cite{grauman2022ego4d}.
Ego4D is a large scale egocentric video dataset contains various daily activities. Each Ego4D video has an average duration of 9 minutes. We use the same train-val split as~\cite{carreira2024bl} -- 21.7k original videos for training and 2.3k for validation.

\paragraph{Task Setting.}
We follow the same task setting as~\cite{carreira2024bl} but only change the optimizer.
Specifically, the model takes 4 video frames as input, and is trained to predict another 4 video frames in the future, with a time displacement of 0.16s or 0.64s.
We use the more challenging time displacement of 0.64s.
We experiment on the pixel prediction task on both datasets,~\ie~the model is trained to predict future pixels, in a sequential way.
For evaluation, we monitor both the in-stream and out-of-stream performance introduced in~\cite{carreira2024bl},
in other words, we report the temporally aggregated performance on the training video stream and also the validation video stream.
This setting can be viewed as a test-time adaptation scenario,
that a pretrained model is expected to adapt well on one video stream (in-stream performance), as well as keep its generalizability on other unseen video streams (out-of-stream performance). 

\paragraph{Architecture.}
We use a ViT-L backbone pretrained on ImageNet-21K classification task as in~\cite{carreira2024bl}.
Notice that~\cite{carreira2024bl} also uses a stronger `Guided Future Prediction' pretraining checkpoint which we are not able to reproduce.
The output of the ViT-L backbone is fed to a randomly-initialized linear layer for future pixel prediction task.
The entire model including the pretrained backbone and the linear layer is trained end-to-end.

\vspace{-1mm}
\paragraph{Implementation Details.}
The model is trained on 24h of training video stream at 25fps, 
given that at each training step, the model takes 4 frames as input without overlapping,
which would be 540k training samples ($24\times3600\text{s}\times25\text{fps} / 4$).
Following~\cite{carreira2024bl} that accumulates gradients every 16 training steps,
equivalently we train the model with a batch size of 16, using the `batch-along-time' setting.
All the experiments use a learning rate of $10^{-4}$, and a cosine-decayed learning rate schedule with linear warm-up.
We report pixel mean squared error (MSE) and peak signal-to-noise ratio (PSNR) 
for the future frame prediction task. A lower MSE and a higher PSNR indicate better performance.

\paragraph{Discussion.}
The results are shown in Table~\ref{table:blv1}.
Notice that our reproduction using the same setting as~\cite{carreira2024bl} (ViT-L-I21K-CLS, with RMSProp optimzer) performs better than the the reported results on pixel MSE (0.032 / 0.026 vs. 0.059 / 0.073 on Ego4D in/out-of-stream).
The proposed Orthogonal-AdamW optimizer further surpasses the baseline AdamW and RMSProp optimizer on both Ego4D-Stream and ScanNet-Stream, on both in-stream and out-of-stream performance.
The in-stream improvements observed indicate our Orthogonal-optimizer can be used for other test-time adaptation tasks beyond representation learning.

\section{Conclusion}
\label{sec:conclusion}

We propose a simple geometric modification to standard optimizers that update with \emph{orthogonal} gradients during training,
in order to decorrelate consecutive batches when training from continuous streams of videos.
We demonstrate three training scenarios which operates on sequential videos: 
representation learning from a single long video, representation learning from large-scale multi-video datasets, and the task of future frame prediction.

Our experiments show that the orthogonal optimizer, in particular Orthogonal-AdamW, 
is able to regularize the learning process and obtain better performance than baseline optimizers for all three tasks.

\section*{Acknowledgement}
We thank James Martens for technical suggestions on the optimizers, and Jean-Baptiste Alayrac and Matthew Grimes for reviewing the manuscript.
We also thank Carl Doersch, Ignacio Rocco, Michael King, Yi Yang and Yusuf Aytar for helpful discussions.

{
    \small
    \bibliographystyle{ieeenat_fullname}
    \bibliography{main}
}

\clearpage
\setcounter{page}{1}
\maketitlesupplementary

\noindent This document provides additional materials 
including implementation details, analysis, ablation studies, 
and additional results that support the main paper.

\section{More Implementation Details}

The implementation details of three scenarios from the main paper Section 4
are shown in Table~\ref{tab:implementation_dora},
Table~\ref{tab:implementation_videomae} and Table~\ref{tab:implementation_bl}.

\begin{table}[h]
\setlength{\tabcolsep}{2pt}
\small
    \centering
    \begin{tabular}{l|ll}\toprule
              & DoRA Pretrain & Linear probe      \\ \hline
        architecture    & ViT-S/16 & ViT-S/16 \\
        embedding dim   & 384  & 384      \\ 
        \# heads & 6    & 6  \\
        \# blocks & 12  & 12 \\ 
        encoder out\_dim & 65536 & N/A \\ \hline 
        
        dataset          & $\text{WT}_\text{venice}$ & ImageNet \\
        \# local crops  & 6             & N/A         \\
        \# global crops & 2             & N/A         \\
        \# input frames  & 8             & N/A          \\
        input fps        & 1             & N/A   \\  
        input resolution & $224\times224$ & $224\times224$ \\
        learning rate    & 0.0005        & 0.01       \\ 
        lr schedule      & Warmup + Cosine & Warmup + Cosine \\
        optimizer        & N/A (varied)   & SGD, m=0.9  \\ 
        weight decay     & $0.04\rightarrow0.4$  & 0           \\ 
        learnable param  & all           & last layer \\
        total batch size & 32 clips      & 512 images \\
        \# epochs        & 1             & 100        \\
        \bottomrule
    \end{tabular}
    \caption{Implementation details of the DoRA experiments 
    in the main paper Section 4.1.}
    \label{tab:implementation_dora}
\end{table}

\begin{table}[h]
\setlength{\tabcolsep}{2pt}
\small
    \centering
    \begin{tabular}{l|ll}\toprule
              & VideoMAE Pretrain  & Linear/Attn probe      \\ \hline
        architecture    & ViT-B/16 & ViT-B/16 \\
        embedding dim   & 768  & 768      \\ 
        \# heads & 12    & 12  \\
        \# blocks & 12  & 12 \\ \hline 
        
        dataset          & K400 / SSV2   & SSV2 \\
        mask ratio       & 0.9           & N/A  \\
        \# input frames  & 16            & 16          \\
        input fps        & 12            & 12   \\  
        input resolution & $224\times224$ & $224\times224$ \\
        learning rate    & 0.0003        & 0.0003       \\ 
        lr schedule      & Warmup + Cosine & Warmup + Cosine \\
        optimizer        & N/A (varied)   & AdamW  \\ 
        weight decay     & 0.05          & 0           \\ 
        learnable param  & all           & linear layer / attn block \\
        total batch size & 512 clips     & 32 clips \\
        \# iterations    & 260k          & 40k        \\
        \bottomrule
    \end{tabular}
    \caption{Implementation details of the VideoMAE experiments 
    in the main paper Section 4.2.}
    \label{tab:implementation_videomae}
\end{table}

\begin{table}[h]
\setlength{\tabcolsep}{2pt}
\small
    \centering
    \begin{tabular}{l|l}\toprule
              & Future prediction  \\ \hline
        architecture    & ViT-L/16 \\
        embedding dim   & 1024      \\ 
        \# heads & 16     \\
        \# blocks & 24  \\ \hline 
        
        dataset          & Ego4d / ScanNet  \\
        \# input frames  & 4                \\
        input fps        & 30             \\  
        input resolution & $224\times224$  \\
        \# output frames  & 4                \\
        prediction $\Delta t$  & 0.64s                \\
        learning rate    & 0.0001              \\ 
        lr schedule      & Warmup + Cosine  \\
        optimizer        & N/A (varied)    \\ 
        weight decay     & $1\times10^{-5}$                  \\ 
        learnable param  & all            \\
        \# steps per update & 16         \\
        \# iterations       & 540k       \\
        \bottomrule
    \end{tabular}
    \caption{Implementation details of the Future prediction experiments 
    in the main paper Section 4.3.}
    \label{tab:implementation_bl}
\end{table}

\section{More Analysis on the Orthogonal Optimizer}

This section provides more analysis to 
have a deep understanding about the orthogonal optimizer, 
in particular Orthogonal-AdamW.

\paragraph{Alternative Option: Downscale Learning Rate.}
Based on the main paper Figure 2 and Equation 4,
readers might question whether the orthogonal optimizer is
effectively using a smaller learning rate.
We experiment with an alternative design choice that indeed 
reduces the learning rate based on gradient similarity.
For example, one can re-scale the learning rate
based on the similarity between the current and the previous gradient.
Formally it can be written as,
\begin{equation}
\begin{aligned}
\lambda &= 1 - \cos{(g_t, g_{t-1})} \in [0, 2] \\
\theta_t &= \theta_{t-1} - \lambda \eta g_t
\end{aligned}\label{eq:downscale}
\end{equation}
where $\eta$ is the learning rate and $\lambda$ is the gradient multiplier.
From the practical observation (\eg~the main paper Figure 3),
we notice that $\cos{(g_t, g_{t-1})}$ is mostly positive, therefore the learning rate
multiplier $\lambda$ mostly has a value within $[0, 1]$, having an effect of reducing the learning rate.

\begin{table}[h]
\setlength{\tabcolsep}{4pt}
\small
    \centering
    \begin{tabular}{l|ll}
    \toprule
               & \multicolumn{2}{c}{Ego4D: MSE$\downarrow$ / PSNR$\uparrow$} \\ \hline
        optimizer         & in-s.         & out-of-s. \\ \hline
        AdamW             & 0.034 / 15.9  & 0.026 / 16.8  \\
        Slower-AdamW      & 0.033 / 16.0  & 0.025 / 17.1  \\ 
        Orthogonal-AdamW  & \textbf{0.031 / 16.4}  & \textbf{0.023 / 17.6}  \\
    \bottomrule
    \end{tabular}
    \caption{Additional results on the future prediction task. The `in-s.' and `out-of-s.' denote in-stream results and out-of-stream results respectively, 
    as same as the main paper Table 3.}
    \label{tab:appendix_bl}
\end{table}

We apply the learning rate scaling method in Eq~\ref{eq:downscale} to the AdamW optimizer,
and name this variant as `Slower-AdamW'.
The experimental results are shown in Table~\ref{tab:appendix_bl}.
The results show that the proposed Orthogonal-AdamW clearly outperform Slow-AdamW on both the in-stream and out-of-stream settings.
Reducing learning rate as in `Slower-AdamW' would avoid over-optimizing along one gradient direction, 
but it is insufficient to actually learn the new signals from correlated gradients.
This result highlights that our method is different from only changing the learning rate 
based on the similarity between consecutive gradients. 

\begin{table}[h]
\setlength{\tabcolsep}{2pt}
    \small
    \centering
    \resizebox{\linewidth}{!}{
    \begin{tabular}{lll|ll}
    \toprule
        seq. video processing   & BS  & optimizer & LP $\uparrow$ & Attn $\uparrow$ \\ \hline
        batch-along-\textbf{time}  & 512 & AdamW            & 16.4              & 46.1 \\
        batch-along-\textbf{time}  & 512 & Orthogonal-AdamW & \textbf{18.4}     & \textbf{48.0} \\ \hline 
        batch-along-\textbf{time}  & 256 & AdamW            & 19.0              & \textbf{47.8} \\
        batch-along-\textbf{time}  & 256 & Orthogonal-AdamW & \textbf{19.9}     & 47.7 \\ \hline 
        batch-along-\textbf{time}  & 128 & AdamW            & \textbf{20.0}     & \textbf{49.2} \\
        batch-along-\textbf{time}  & 128 & Orthogonal-AdamW & 18.7              & 47.9 \\ \hline\hline  
        
        batch-along-\textbf{video} & 512 & AdamW            & 9.5                & 30.3 \\
        batch-along-\textbf{video} & 512 & Orthogonal-AdamW & \textbf{10.4}      & \textbf{32.6} \\ \hline 
        batch-along-\textbf{video} & 256 & AdamW            & 8.3                & 25.7 \\
        batch-along-\textbf{video} & 256 & Orthogonal-AdamW & \textbf{13.2}      & \textbf{37.6} \\ \hline 
        batch-along-\textbf{video} & 128 & AdamW            & 17.1               & 41.6 \\
        batch-along-\textbf{video} & 128 & Orthogonal-AdamW & \textbf{18.3}      & \textbf{44.0} \\
    \bottomrule
    \end{tabular}}
    \caption{Effect of batch size on VideoMAE pretrained on SSV2 and evaluated on SSV2,
    using the same setting as the main paper Table 2. The `BS' denotes Batch Size.}
    \label{tab:appendix_videomae}
\end{table}

\paragraph{Impact of the Batch Size.}
The calculation of orthogonal gradients highly depends on the size of the mini batch.
We analyze the impact of batch size on VideoMAE pretraining task on SSV2. The experimental results are shown in Table~\ref{tab:appendix_videomae}.
Note that when reducing the batch size, 
we proportionally increase the number of training iterations to ensure each experiment is trained on the same number of samples.
For example, comparing with $\text{BS}=512$, 
the experiments using $\text{BS}=256$ and $\text{BS}=128$ are trained with $2\times$ and $4\times$ longer training schedules.

The experimental results show a few interesting trends.
First, the absolute performance of `batch-along-time' strategy does not change much with different batch sizes, and the Orthogonal-AdamW outperforms AdamW on larger batch sizes (512, 256), but underperforms AdamW on smaller batch size (128).
Second, the VideoMAE trained with `batch-along-time' strategy generally performs better with smaller batch size, and the Orthogonal-AdamW clearly outperforms AdamW on this setting.

\paragraph{Impact of the Momentum Parameter.}
In the main paper Equation 3, we introduce a hyper-parameter $\beta$
controlling the update rate of the momentum.
We experiment with different $\beta$ values in Table~\ref{tab:appendix_beta}.
Generally, a large value of $\beta$ (close to 1.0) leads to a `smoother' momentum value;
a lower value of $\beta$ (close to 0.0) makes the momentum more fluctuate and less robust to noise, as the current value has large impact to the momentum.
At the extreme case when $\beta=0$, it means the momentum is not used. In our case, it means the orthogonal gradient is computed w.r.t. the previous gradient.
The results in Table~\ref{tab:appendix_beta} shows that $\beta \geq 0.9$ works well,
and there is almost no difference using $0.9$ or $0.99$.
By default, we use $\beta=0.9$ in the main paper experiments.

\begin{table}[h]
\setlength{\tabcolsep}{5pt}
\small
    \centering
    \begin{tabular}{l|ll}
    \toprule
         Orthogonal-AdamW  & \multicolumn{2}{c}{Ego4D: PSNR$\uparrow$} \\ \hline
         $\beta$           & in-s.        & out-of-s.      \\ \hline
         0                 & 16.1         & 17.1               \\
         0.5               & 16.3         & 17.4               \\
         0.9               & \textbf{16.4}         & \textbf{17.6}               \\
         0.99              & \textbf{16.4}         & \textbf{17.6}               \\
    \bottomrule
    \end{tabular}
    \caption{Impact of the momentum parameter in Orthogonal-AdamW.
    This is a future prediction task on Ego4D-Stream, as same as the main paper Table 3.}
    \label{tab:appendix_beta}
\end{table}

\begin{table}[h]
\setlength{\tabcolsep}{2pt}
\small
    \centering
    \begin{tabular}{l|l}
    \toprule
        optimizer & ImageNet top1$\uparrow$ \\ \hline
        \color{gray}AdamW~\cite{dosovitskiy2020vit} & \color{gray}77.9 \\
        AdamW (repro)            & 77.8     \\
        Orthogonal-AdamW         & 76.5     \\
    \bottomrule
    \end{tabular}
    \caption{ImageNet classification results with ViT-B/16 architecture. 
    The models are trained from scratch following the recipe in the original ViT paper~\cite{dosovitskiy2020vit}. 
    Note that the first row is the official ViT result from~\cite{dosovitskiy2020vit}.
    `repro' means our reproduction.
    }
    \label{tab:appendix_imagenet}
\end{table}

\section{Does it Work on ImageNet Classification?}
In the main paper Section 4,
we have shown the Orthogonal-AdamW outperforms AdamW on various 
self-supervised video learning scenarios, even on \emph{shuffled} video clips (VideoMAE results in the main paper Table 2).
Naturally, we would like to know if the orthogonal optimizer can be applied to 
general \emph{supervised learning} tasks.
In this section, we compare Orthogonal-AdamW with AdamW on the classic ImageNet classification task.
Results are shown in Table~\ref{tab:appendix_imagenet}.
We use a ViT-B/16 architecture and follow the training recipe from~\cite{dosovitskiy2020vit}.
First, our reproduction matches the reported ViT performance on ImageNet (77.8 vs 77.9).
Second, we find the Orthogonal-AdamW underperforms AdamW by 1.3\% on this task.
It is probably because ImageNet mini-batches follow IID distributions more closely, 
and the gradients from consecutive batches have negligible correlation.
In this case, optimizing the orthogonal gradients does not bring informative learning signals.

\end{document}